# Development of control algorithms for mobile robotics focused on their potential use for FPGA-based robots

Andrés-David Suárez-Gómez,[1] ANDRES A. HERNANDEZ ORTEGA,[1,2]

[1] UNIVERSIDAD NACIONAL ABIERTA Y A DISTANCIA – Grupo de Investigación Byte in Design
[2] UNIVERSIDAD PEDAGOGICA Y TECNOLOGICA DE COLOMBIA.

## Abstract

This paper investigates the development and optimization of control algorithms for mobile robotics, with a keen focus on their implementation in Field-Programmable Gate Arrays (FPGAs). It delves into both classical control approaches such as PID and modern techniques including deep learning, addressing their application in sectors ranging from industrial automation to medical care. The study highlights the practical challenges and advancements in embedding these algorithms into FPGAs, which offer significant benefits for mobile robotics due to their high-speed processing and parallel computation capabilities. Through an analysis of various control strategies, the paper showcases the improvements in robot performance, particularly in navigation and obstacle avoidance. It emphasizes the critical role of FPGAs in enhancing the efficiency and adaptability of control algorithms in dynamic environments. Additionally, the research discusses the difficulties in benchmarking and evaluating the performance of these algorithms in real-world applications, suggesting a need for standardized evaluation criteria. The contribution of this work lies in its comprehensive examination of control algorithms' potential in FPGA-based mobile robotics, offering insights into future research directions for improving robotic autonomy and operational efficiency.

## Introducción

The design and optimization of control algorithms is often investigated to improve different aspects of a robot's performance, such as path-following control for moving from one point to another. There is a large amount of research on robot control algorithms and new approaches are constantly being proposed. However, the problem arises of the difficulty in comparing the results of these published research and evaluating the quality of the studies. Furthermore, in robotics publications, performance evaluation criteria are often overlooked, making it difficult to make an objective comparison of algorithms. Tests, whether in simulation or experimental, are often limited to measuring the length of the path traversed or the time it takes the robot to complete a task. In addition, there are few standard methods to evaluate the capabilities and limitations of these systems in a comparable manner (Norton et al, 2019).

Research in this field is typically conducted in controlled laboratory settings to validate proofs of concept and establish useful comparisons. However, it is important to keep in mind that the results obtained may differ to some extent from the actual operation of a robot, as the latter is characterized by the presence of uncertainty (Martins et al., 2020). In addition, there is a lack of consensus on performance evaluation criteria, which often vary from one study to another. This lack of consensus makes it difficult to compare the capabilities of navigation algorithms and detracts from the rigor of the evaluation of advances in this field. Consequently, a comprehensive evaluation system is lacking (Ren et al., 2020).

# Backgroung

**Differential drive mobile robots:** Differential drive mobile robots are a common category in mobile robotics. These robots are characterized by having two independently driven wheels, giving them great maneuverability and mechanical simplicity. Their ability to change the speed and direction of each wheel independently allows for precise movements and turns on the spot. A seminal text that addresses these types of robots is "Introduction to Autonomous Robots" by Correll and Teller (2011). Previous research has explored specific control algorithms for these types of robots, highlighting the importance of differential kinematics in path planning.

**FPGA technology in robotics:** FPGA technology plays a crucial role in robotics, especially in the efficient implementation of control algorithms. FPGAs provide parallel processing and reconfiguration capabilities, which is particularly beneficial in dynamic robotic environments. To further understand the application of FPGAs in robotics, "Digital Design and Computer Architecture" by Harris and Harris (2010) provides a detailed overview of FPGA architecture and application. Previous research has demonstrated the successful use of FPGAs in the implementation of embedded systems for real-time control of robots.

**Control algorithms in robotics:** Control algorithms are essential for the effective operation of robots in diverse environments. "Principles of Robot Motion: Theory, Algorithms, and Implementations" by Choset et al. (2005) is a comprehensive reference that addresses various control algorithms used in mobile robotics. From path planning algorithms to feedback control techniques, the existing literature highlights the need for adaptive and efficient algorithms. Previous research has explored the application of techniques such as proportional-integral-derivative (PID) control in stabilization and tracking of mobile robots.

**Previous research and contributions:** Review of previous research reveals that the field of mobile robotics has experienced significant advances in control algorithms, application of FPGA technologies, and differential drive robot design. For example, "Robotics Research: The 13th International Symposium ISRR" (2017) presents several contributions in the field of autonomous navigation and control of mobile robots. Previous contributions have demonstrated the effectiveness of heuristic algorithm-based path planning and implementation of real-time control systems using FPGA in challenging robotic environments.

# Hardware Architecture

**Role of the FPGA in the robot:**

FPGA plays a crucial role in the design and implementation of robotic systems, providing significant flexibility and efficiency. The reconfigurability of FPGAs allows them to adapt to changes in the operating environment and control requirements of the robot. This feature is especially valuable in applications where the complexity of the control algorithm can vary dynamically.

In the specific context of a robot, the FPGA can be used to implement real-time control algorithms, improving the responsiveness and accuracy of the system. In addition, the parallel architecture of FPGAs is ideal for executing multiple tasks simultaneously, which can be essential in robotic applications that require fast sensory data processing and real-time decision making.

A highlight of the FPGA's role in robots is its ability to improve energy efficiency. The implementation of control algorithms in configurable hardware allows the computational load on the robot's main CPU to be reduced, which can translate into lower power consumption and increased autonomy.

Previous research has successfully explored the integration of FPGAs into task-specific robots, demonstrating remarkable improvements in processing speed and system efficiency. Specialized literature, such as "FPGA-Based Implementation of Signal Processing Systems" by Woods et al. (2008), provides a deeper understanding of how FPGA can contribute to the overall performance of a robot.

## FPGA Implementation

**Basic Concepts:**

**FPGA (Field-Programmable Gate Array):** An FPGA is a digitally configurable semiconductor device that allows designers to program or configure the digital logic and interconnect within the chip according to their specific needs.

**Programmable Logic Array (PL):** The programmable logic array is the core of an FPGA and consists of an arrangement of logic blocks and connections that can be programmed to perform specific tasks.

**Logic Blocks:** Logic blocks are basic units that contain logic gates and flip-flops. These blocks are configured to implement desired logic functions.

**Interconnect:** The interconnect consists of the paths that connect the logic blocks and other elements within the FPGA, enabling communication and data transfer.

**Configuration:** Programming an FPGA involves establishing the functionality of its logic blocks and interconnect using a configuration file. This process determines how the FPGA will perform specific tasks.

**FPGA Advantages:**

**Flexibility and Reconfigurability:** The main advantage of FPGAs is their ability to be reconfigured according to the designer's needs. This allows the hardware to be adapted to different applications without the need to change the physical components.

**High Performance:** FPGAs can offer significant performance compared to other programmable technologies, especially in applications requiring parallelism and real-time data processing.

**Rapid Development:** Reconfigurability enables rapid prototyping and proof-of-concept development. Designers can easily make iterations and adjustments without having to redesign the hardware.

**Custom Parallelism:** The parallel architecture of FPGAs enables highly parallel algorithms to be implemented, improving efficiency and processing speed compared to sequential implementations.

**Energy Efficiency:** By enabling the specific implementation of the hardware required for a task, FPGAs can be more energy efficient compared to general-purpose solutions, reducing power consumption in embedded systems and mobile devices.

**Adaptability to Change:** The ability to reconfigure makes it easier to adapt to changes in design requirements, which is especially valuable in environments where requirements may evolve over time.

**Implementation of Specific Functions:** FPGAs are ideal for implementing specific functions efficiently, such as signal processing algorithms, image processing, and embedded system control.

## Materials and Methods

Normally controlling the inputs of a monocycle model is about applying the traditional feedback PID controller and selecting the appropriate input u = $(v\omega)^T$, given by eq:

$$U(t) = PID(e) = K_p e(t) + K_I \int_0^1 e(\tau)\, d(\tau) + K_D \frac{de\,(t)}{dt} \quad (1)$$

In the context of each task detailed below, the term 'e' refers to the error between the desired value and the value obtained as a result. The constants Kp, Kl and KD represent the proportional, integrative, and derivative gains respectively, while 't' refers to time. The control gains used in this study are determined by adjusting different values in order to achieve satisfactory responses. If the vehicle travels at a constant speed, v = v0, then the control input will only change along with the angular velocity, ω, in this manner:

$$w = PID(e) \quad (2)$$

## Results and Discussion

**Space-time Criteria**

Performance criteria related to space-time dimensions are widely used and allow a quantitative evaluation and comparison of results in real experiments or in simulation. In the article by Munoz et al. (2014), several typical criteria in navigation and obstacle avoidance are described, such as mission success, robustness in narrow spaces, path length, time required to complete the mission, control periods, average distance to target, distance to obstacles, and smoothness of trajectory, among others.

Among these criteria, those related to space-time dimensions are the simplest and most used. An optimal trajectory from the point of view of reaching the goal is one that follows a straight line of the minimum possible length and without curvature between the starting point (xi, yi) and the arrival point (xn, yn), traversed in the shortest time. This approach assumes linearity and constant velocity of the robot in its trajectory towards the goal (Munoz-Ceballos, N. D. et al., 2022).

The Dynamic Window Approach (DWA) is a well-known collision avoidance navigation algorithm. It was initially proposed by Dieter Fox and his team. DWA works in real time and reacts to changing situations as they occur. In recent years, the DWA cost function has undergone several extensions and improvements. This approach determines safe and optimal translation (v) and rotation (w) speeds directly by creating velocity profiles that take into account robot dynamics and speed and acceleration constraints. The search for suitable velocities mainly involves three subspaces, including the space of possible values of v (Mohammadpour et al, 2021).

The space of possible velocities is divided into three subspaces according to the kinematic constraints of the robot:

Space of Possible Velocities (Vs): This subspace considers the kinematic constraints of the robot and represents all velocities that are physically possible for the robot based on its mechanical characteristics.

Admissible Velocity Space (Va): This subspace considers the velocities that allow the robot to stop without colliding with an obstacle. These velocities are restricted by the robot's ability to stop safely and without collision.

Space of Possible Speeds with Consideration of Robot Acceleration Limitations (Vd): This subspace considers the acceleration limitations of the robot. It represents the possible velocities considering the robot's limited ability to change its velocity quickly due to its acceleration constraints.

$$Vr = Vs \cap Va \cap Vd \quad (3)$$

Where Vr is the optimal velocity search space, which is selected by maximizing the following objective function:

$$G(v,w) = \propto * h(v,w) + \beta * d(v,w) + \gamma * v_F(v,w) \quad (4)$$

Where "h" measures the alignment of the robot with the target direction, "d" represents the distance to the nearest obstacle, and "vF" is the forward velocity of the robot. The values of α, β and γ are adjustable constants that determine the relative importance of these three measures in the objective function. In summary, the DWA method generates numerous possible local online trajectories and then selects the most suitable one based on the objective function. Finally, the most suitable velocity is run to follow the selected local trajectory.

**Success in reaching the goal**

This criterion is generally given in terms of percentage (%), it consists of counting the percentage of success in completing a navigation mission with respect to the total number of attempts. Some researchers also set a time limit, but sufficient to successfully complete the given navigation mission, this in order to rule out tests where the robot gets stuck navigating in an endless loop (McGuire et al, 2019).An additional challenge for control algorithms is their performance in navigating through narrow passages or corridors, therefore, an additional criterion to consider may be the robustness in narrow spaces: number of narrow passages successfully traversed.

As an example it is used in navigation robots that must find the exit to a maze, in such situation the triangulation of the position is done by the configuration of the solved triangle, where we want to calculate the height (h) of the triangle with sides 'a' and 'b' and angle β. 'c' is the side of the triangle that will be unknown, so a formula will be developed that will only use 'a,' 'b' and β.

$$A = \frac{c.h}{2} \quad (5)$$

$$A = \frac{a.b.\sin\beta}{2} \quad (6)$$

The cosine theorem is applied to triangulate the height and develop the triangle by replacing the above equations.

$$h = \frac{a.b.\sin\beta}{\sqrt{a^2 + b^2 - 2.a\cos\beta}} \quad (7)$$

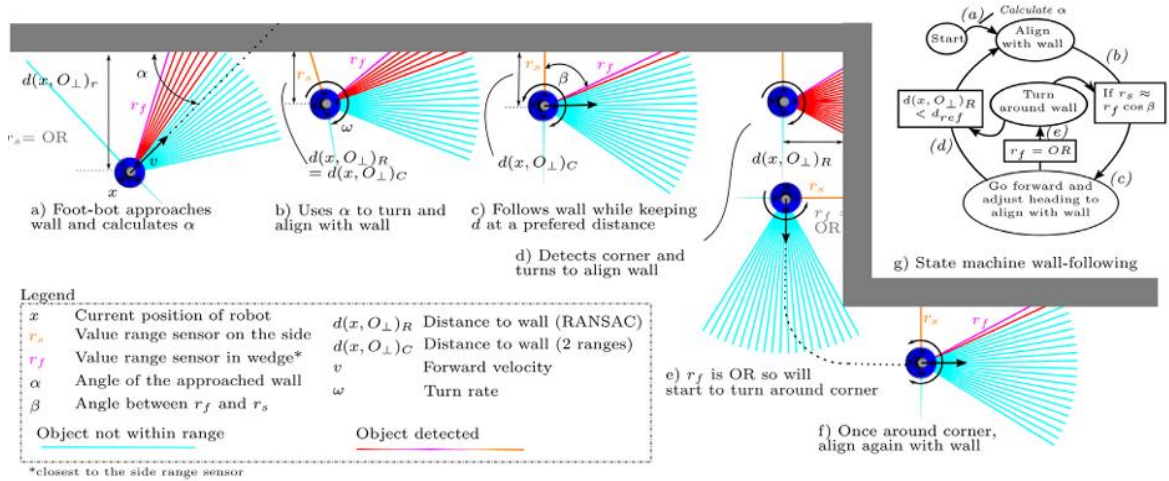

Figure 1. Diagram to explain the wall-tracking paradigm for a local address on the right-hand side with "(g)" representing the corresponding state machine. "OR" stands for "out of range". Taken from (McGuire et al, 2019).

**Time reaching the target**

Execution time is a fundamental criterion used in most articles comparing algorithms. It consists of measuring the time the robot takes to reach the goal. In simulations of deterministic systems, the same result is obtained under the same simulation conditions. However, to bring the simulation closer to reality, where there are factors such as battery wear, friction between the wheels and the ground, environmental conditions, among others, noise can be introduced into the system. This is achieved by adding disturbances in the sensor readings or in the control signals to the actuators. In simulation, random number generators can be used to model the error or noise.

Another related criterion is the number of processing cycles, which is equivalent to the approximate number of operations performed to complete a mission. It is important to note that different types of robots may have different types of processors, i.e., different computational capabilities. This influences the total time the robot takes to complete the mission and the comparison of algorithms using this criterion (Tai et al., 2020).

**Control periods**

Control periods refer to the number of times the planner makes decisions to reach the goal. This measure is related to the number of iterations or steps the robot needs to complete the mission. If the robot moves at a constant linear velocity (v), the control periods provide an estimate of the time taken to complete the mission. The greater the number of control periods, the greater the number of decisions made and, potentially, the greater the time required to reach the goal.

**Error-based Criteria**

In a control system, error is defined as the difference between the controlled variable (also known as the process variable) and the reference value or set-point. In a control system, the objective is that the error tends to zero, indicating good system performance.

One way to evaluate the performance of a control system is to quantify the cumulative error. In the case of a mobile robot, the cumulative error provides a numerical measure of how "good" the performance of a specific controller is for the robot's drive or steering system.

In discrete-time controllers, it is necessary to know the error e(nT) at each sampling instant, where T is the sampling period and fs is the sampling frequency. The sampling frequency has a direct impact on the accuracy of the measurements and the ability of the controller to detect and correct the error over time.

Error-based or error-integral performance criteria have a well-established theoretical foundation in control systems, and are widely used to evaluate and improve controller performance (Domański, 2020). These criteria allow the system behavior to be analyzed in terms of error reduction and control stability.

**Final error**

Final error refers to the discrepancy between the final position of the vehicle and the end point of an established reference trajectory. It is calculated by measuring the distance between the final position of the vehicle and the desired end point of the reference trajectory. This measurement is especially useful in robotic underwater vehicles, as it allows detecting if the vehicle has lost the trajectory halfway through the tracking or if it has reached an incorrect end position (Perez et al., 2018).

As an example, the article mentions that by means of a developed platform, it is possible to automatically evaluate the performance of the obtained solutions using a specific metric. In this context, a typical metric in control algorithms for trajectory tracking is used: the integrated squared error (ISE) and the final error. The ISE is calculated as the sum of the distances to the ideal trajectory, which is two meters above the pipe, over time. This measure is inversely related to tracking quality, as it considers both time-spent and tracking accuracy.

The rate of increase is significantly higher as the distance between the vehicle and the optimal trajectory increases, but it also penalizes excessively slow trajectory tracking. In addition, it is essential to determine whether the end of the pipeline has been reached. To achieve this, the distance between the end position of the vehicle and the end of the pipeline is calculated. In this way, it is possible to identify if the vehicle has lost the pipeline during the middle of the tracking or if it has incorrectly detected the end. Taking these measurements into consideration, the final evaluation is determined by equation 7.

$$evaluation = (1 - error^2) * (0{,}1 - \frac{meanerror^2}{0{,}1} + \frac{time - ref}{100} \quad (7)$$

Where "error" represents the final error, "mean error" is the average error over the tracking calculated from the ISE, "time" indicates the time required to perform the intervention, and "ref" refers to the time reference for each scenario, determined as a function of distance traveled, turns and changes in altitude. The first term of the equation evaluates the final position of the vehicle, penalizing those situations in which the vehicle is far from the target. The second term evaluates the average error along the trajectory. The last term is a bonus that favors fast tracking and penalizes slow tracking based on the complexity of the pipeline path.

In general controller design, there are commonly used performance criteria, such as indices involving the integral of the error (Suarin et al., 2019). These criteria are based on cumulative error and can be applied to reference trajectory tracking, indicating the error along the entire path between the reference trajectory and the actual trajectory followed by the robot. These indices are also used in position control, distance, orientation, multiple robot formation, among others (Caruntu et al., 2019) (Farias et al., 2020). The lower the error, the better the trajectory traversed and, consequently, the better the control algorithm.

**Safety Criteria**

Related research addressing performance criteria in robot control and its safety in navigation is described in the study conducted by Marvel and Bostelman in 2014. These performance criteria focus on the safety of the robot while moving along a given trajectory, taking into account factors such as the distance between the vehicle and the obstacles encountered in its path, as well as the number of collisions occurring during navigation (Munoz et al., 2014). These investigations seek to ensure safe travel and avoid possible accidents or damage during the operation of the robot.

The average distance to obstacles during the entire navigation mission is another performance criterion used in robot control. This criterion makes it possible to evaluate how close or far the robot is from obstacles in its environment during the entire trajectory followed. In an unobstructed environment, this maximum value will be higher, since the robot can move freely without restrictions. On the other hand, if the average distance to obstacles index deviates less from the maximum value, it means that the path followed by the robot passed through more obstacle-free zones, indicating better performance in terms of avoidance and safe navigation. This criterion is important to avoid collisions and ensure a clear and obstruction-free path for the robot during its navigation mission.

The minimum average distance to obstacles is another performance criterion used in robot control to evaluate safety during a navigation mission. It averages the minimum distance value measured by each of the robot's n sensors. This criterion provides an idea of the risk taken during the mission in terms of proximity to obstacles.

In unobstructed environments, where there are no obstacles close to the robot, it will be fulfilled that the minimum average distance to obstacles will be equal for all sensors. This indicates that the robot has maintained a safe distance and has not been exposed to significant risk of collision or contact with obstacles.

In contrast, in environments with obstacles, the smaller the minimum mean distance to obstacles, the greater the risk that has been taken during the mission, since the robot has been closer to the obstacles. Therefore, a lower value of this criterion indicates a higher probability of collisions or contact with obstacles.

**Energy Consumption**

Energy consumption is a key aspect in mobile robot applications, as it influences the autonomy of the robot, that is, the time it can function optimally before running out of energy. In recent years, special attention has been paid to this topic (Stefek et al, 2020). If a robot does not meet energy consumption requirements, such as the ability to operate independently and the ability to recharge, its performance, operating time and autonomy are limited (Heikkinen et al, 2018).

Mobile robots rely heavily on batteries as a power source, but these have limited energy capacity. As a result, the robot's operating time is often short, which may be insufficient for tasks or missions that require more time and energy to complete. Increasing run time by using more batteries or directing the robot to a charging station can increase the cost or size of the system, which can cause control issues. An alternative is to improve the energy efficiency of the robot, reducing its energy consumption (Armah et al, 2016).